\pdfoutput=1

\documentclass[11pt]{article}

\usepackage[final]{acl}
\usepackage{comment}
\usepackage{times}
\usepackage{latexsym}
\usepackage{tabularx}
\usepackage{makecell} 
\usepackage{multirow}
\usepackage{algorithm}
\usepackage{algorithmic}

\usepackage{amsmath,amssymb,amsfonts}
\usepackage{colortbl}
\usepackage[T1]{fontenc}

\usepackage[utf8]{inputenc}

\usepackage{microtype}

\usepackage{inconsolata}

\usepackage{booktabs}       
\usepackage{amsfonts}       
\usepackage{nicefrac}       
\usepackage{graphicx}      
\usepackage{graphicx}
\usepackage{multirow}
\usepackage{subfig}
\usepackage{lineno}
\usepackage{pifont} 
\usepackage{hyperref}
\usepackage{color}
\usepackage{colortbl}
\usepackage{graphicx}
\usepackage{booktabs}
\usepackage{multirow}

\newcommand{\checkmarkgreen}{\textcolor{green}{\checkmark}}%
\newcommand{\xmarkred}{\textcolor{red}{\text{\sffamily X}}}%
\title{Flora: Effortless Context Construction to Arbitrary Length and Scale}

\author{
  \textbf{Tianxiang Chen\textsuperscript{1,2,3}},
  \textbf{Zhentao Tan\textsuperscript{4}},
  \textbf{Xiaofan Bo\textsuperscript{4}},
  \textbf{Yue Wu\textsuperscript{5}},
  \textbf{Tao Gong\textsuperscript{1,2,3}}\\
  \textbf{Qi Chu\textsuperscript{1,2,3}},
  \textbf{Jieping Ye\textsuperscript{5}},
  \textbf{Nenghai Yu \textsuperscript{1,2,3}}
\\
  \textsuperscript{1}School of Cyber Science and Technology, University of Science and Technology of China\\
  \textsuperscript{2}Anhui Province Key Laboratory of Digital Security\\
  \textsuperscript{3}CAS Key Laboratory of Electromagnetic Space Information\\
  \textsuperscript{4}Zhejiang Lab\\
  \textsuperscript{5}Independent Researcher
\\
}

\begin{document}
\maketitle
\begin{abstract}
Effectively handling long contexts is challenging for Large Language Models (LLMs) due to the rarity of long texts, high computational demands, and substantial forgetting of short-context abilities. Recent approaches have attempted to construct long contexts for instruction tuning, but these methods often require LLMs or human interventions, which are both costly and limited in length and diversity. Also, the drop in short-context performances of present long-context LLMs remains significant. In this paper, we introduce Flora, an effortless (human/LLM-free) long-context construction strategy. Flora can markedly enhance the long-context performance of LLMs by arbitrarily assembling short instructions based on categories and instructing LLMs to generate responses based on long-context meta-instructions. This enables Flora to produce contexts of arbitrary length and scale with rich diversity, while only slightly compromising short-context performance. Experiments on Llama3-8B-Instruct and QwQ-32B show that LLMs enhanced by Flora excel in three long-context benchmarks while maintaining strong performances in short-context tasks. Our data-construction code is available at \href{https://github.com/txchen-USTC/Flora}{https://github.com/txchen-USTC/Flora}.

\end{abstract}


\section{Introduction}

Large language models (LLMs) are widely used
in many natural language processing tasks. These tasks often require dealing with lengthy text inputs \cite{bai2023longbench,bai2024longbench}, such as long conversation histories \cite{zhong2024memorybank} or long documents \cite{bai2024longbench}. Thus, improving LLMs to handle long-context inputs effectively is critical.

\begin{figure}
    \centering \includegraphics[width=0.48\textwidth,height=0.37\textwidth]{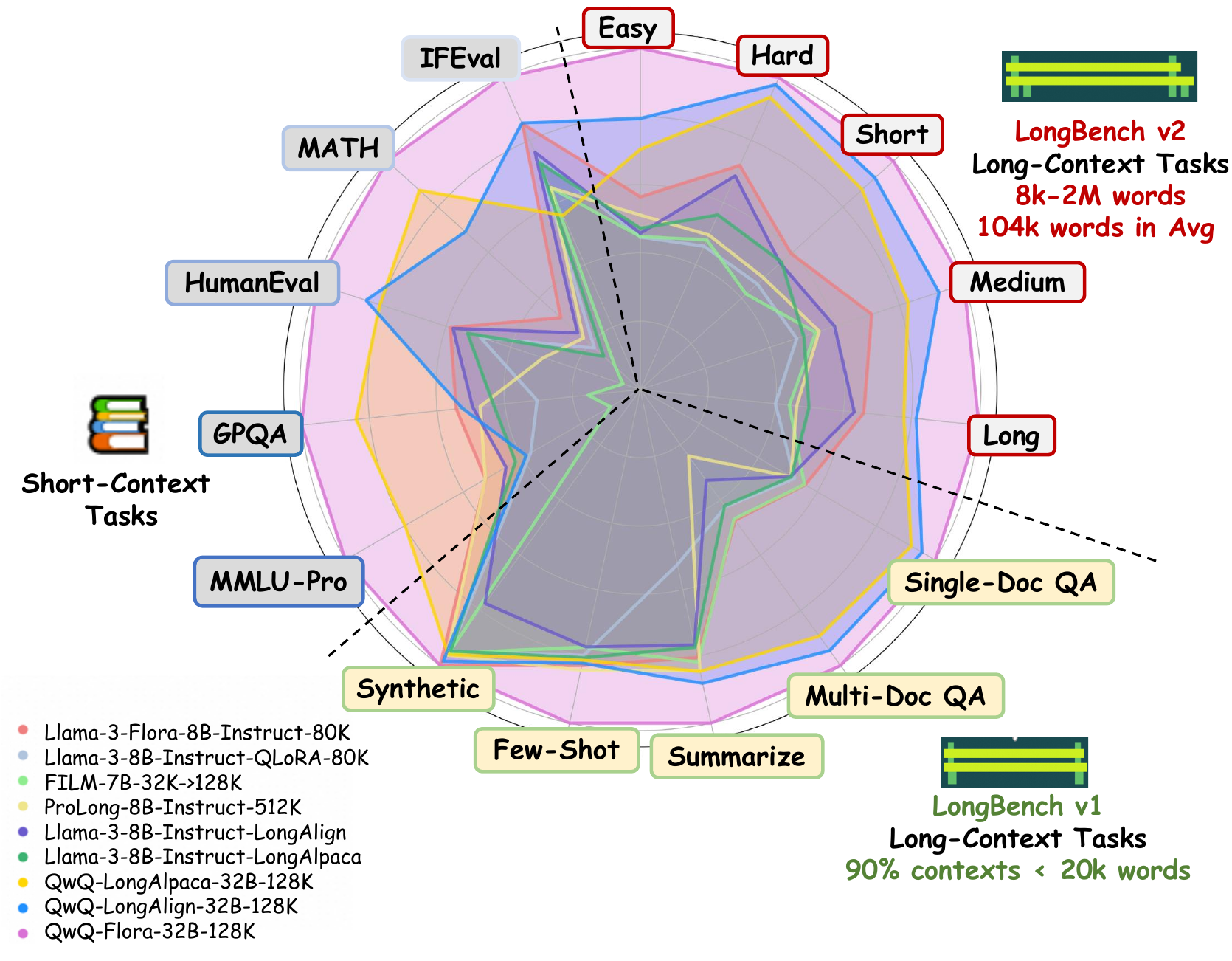}
    \caption{Average scores across long and short context tasks, normalized by the highest score on each task. The scores on LongBench v2 \cite{bai2024longbench} are evaluated in the zero-shot + CoT setting. Our Flora-enhanced models achieve state-of-the-art (SOTA) performances on all long and short context tasks, compared to other models of similar parameter scales.} \label{challenge_radar}
\end{figure}

There are two categories of approaches
to expand LLM's context window. The first focuses on modifying the LLM structure, such as altering the positional encoding \cite{chen2023extending,ding2024longrope} or the attention mechanism \cite{peng2023rwkv,munkhdalai2024leave,gu2023mamba}, and these are referred to as model-level methods. Nonetheless, these methods typically aim to overcome limitations related to models, hardware, etc., in the pursuit of understanding long texts, such as contexts with over 1 million tokens. In addition, data quality and diversity also matter \cite{chen2023longlora} in improving the long-context modeling capabilities of LLMs. Therefore, the second category, data-level methods, focuses on enhancing LLM's long-context capability via long data-constructing aspects \cite{chen2024long,tang2024logo,an2024make,zhang2024extending}. However, as displayed in Table~\ref{method_compare}, most of the current data-level methods necessitate human or LLM involvement in data generation. This approach is expensive and constrained, typically involving datasets of less than 20k samples, each under 10k tokens long. 
Moreover, enhancing long-context performance in LLMs often markedly compromises their short-context capabilities, even though nearly half of their training data is short \cite{chen2023longlora}.

\begin{table}
  \centering
  \resizebox{0.99\columnwidth}{!}{
  \begin{tabular}{c|ccc}
    \toprule
    Challenges&LC Construction Method &Concatenation Method & Flora (Ours) \\
    \midrule
    LLM/Human-free &\xmarkred &\checkmarkgreen & \checkmarkgreen\\
    Low Demand for LC &\xmarkred &\checkmarkgreen &\checkmarkgreen \\
    Length $\&$ Diversity  &\xmarkred &\checkmarkgreen & \checkmarkgreen\\
    Maintain SC Ability  &\xmarkred &\checkmarkgreen &\checkmarkgreen \\
    Length Control  &\checkmarkgreen &\xmarkred &\checkmarkgreen \\
    LC-specific  &\checkmarkgreen &\xmarkred & \checkmarkgreen\\

    \bottomrule
  \end{tabular}}
  \caption{Our Flora addresses the six major challenges of current long-context construction and concatenation strategies. Flora is LLM/human-free, does not require massive long contexts, offers diverse and infinitely lengthy data, preserves LLM's short context abilities, control input and output length and is tailored for long context tasks. "SC" stands for "short context," and "LC" stands for "long context."}
  \label{method_compare}
\end{table}

\begin{figure}
    \centering \includegraphics[width=0.49\textwidth,height=0.18\textwidth]{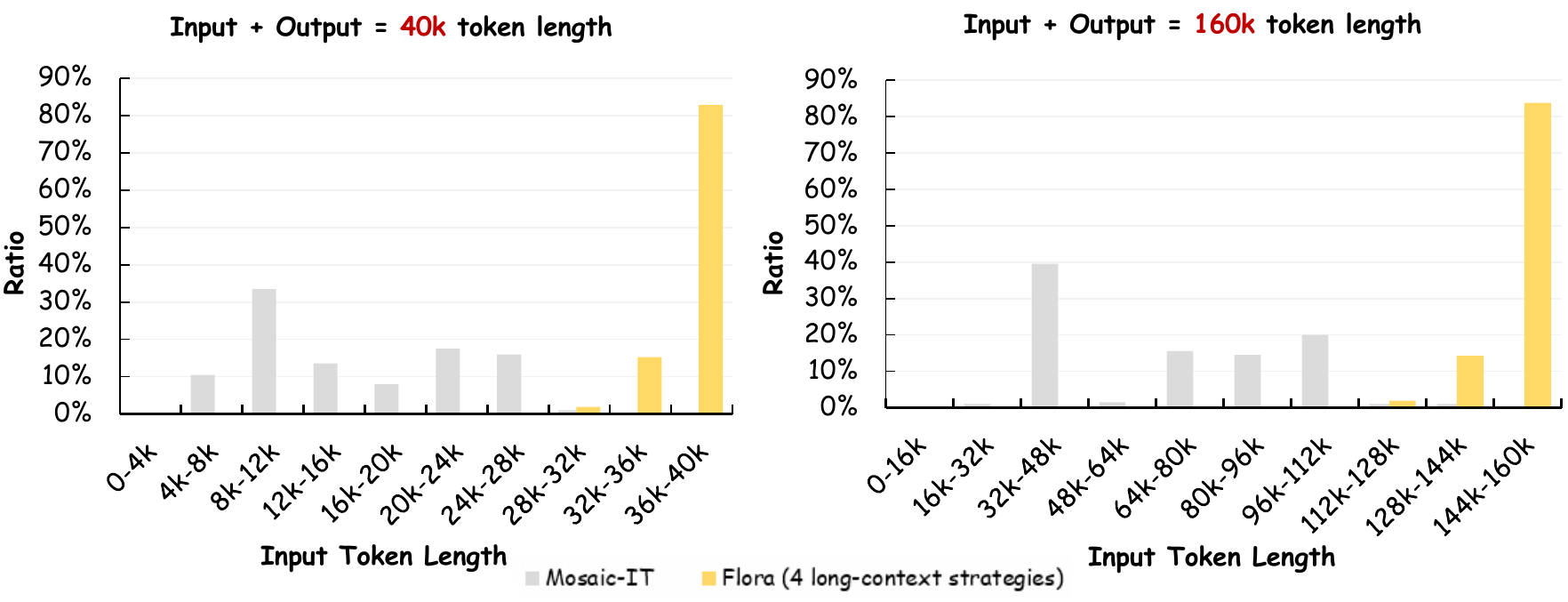}
    \caption{Comparison of output token length distributions between Flora-enhanced and Mosaic-IT enhanced data under fixed total token lengths. The $x$-axis shows the output token length, and the $y$-axis shows the ratio.}\label{compare}
\end{figure}

To overcome the challenges, an intuitive way is to construct long-context data based on short-context data stacking for the following reasons: 1) Compared to the insufficient long-context data, we can easily obtain high-quality open-source general supervised fine-tuning (SFT) datasets of millions or even larger scale \cite{lambert2024t}. 2) Existing pre-trained LLMs are not confused by the concatenated data due to the widespread use of data concatenation technologies in their training stage \cite{wolf-etal-2020-transformers,deepseekai2024deepseekv3technicalreport}. 3) These data naturally ensure the short-context capabilities of fine-tuned models. What's more, Mosaic-IT \cite{li2024mosaic} has proven the potential of data concatenation in  boosting the instruction-following capabilities of LLMs while simultaneously accelerating the training process. However, it cannot regulate the length of inputs and outputs, with outputs generally much longer than the inputs, as shown in Fig.~\ref{compare}. This approach produces suboptimal samples for long-context scenarios and lacks specific designs for enhancing key long-context abilities such as multi-document retrieval and QA.

To take a further step, we propose to concatenate short instructions and their responses as a "long context", and design corresponding instructions to enable the model to answer questions based on a full understanding of this "long context". Following this principle, we introduce Flora, an effortless data construction strategy that can generate instruction-tuning data of any length and scale. Specifically, after obtaining the concatenated long context, we design four instruction templates to simulate common long context understanding tasks such as multi-document question answering and summarization. Furthermore, the responses for the synthesized data are based on the original short context data, eliminating the need for human or LLM intervention. It allows for highly diverse and theoretically limitless-length contexts from existing short instruction tuning datasets with minimal impact on short-context performance and can also flexibly control the input and output length. We curated long-context datasets, Flora-80k and Flora-128k, with maximum token lengths of 80k and 128k. We fine-tuned only 8.5$\%$ parameters of Llama-3-8B-Instruct and 5.5$\%$ of QwQ-32B respectively on these datasets to develop Flora-enhanced models. As displayed in Fig.~\ref{challenge_radar}, our Flora-enhanced models deliver SOTA results on various long-context benchmarks while also significantly excelling in short-context understanding, compared to other models of similar parameter scales.



\section{Related Works} 

\subsection{Model-level Long-Context LLMs} 
Model-level methods mainly adjust model structures. Some of these methods highlight improvements in position encoding design, which enhance the representation of positional information within models \cite{chen2023extending,liu2023scaling,peng2023yarn,ding2024longrope}. Additionally, \cite{bai2024longalign} optimizes batching strategies for efficient data processing. Parameter-efficient training focuses on reducing computational resources while maintaining performance \cite{chen2023longlora}. There are also methods to innovate new model architectures by modifying attention mechanisms \cite{peng2023rwkv,dai2019transformer,munkhdalai2024leave,gu2023mamba}. However, model-level approaches usually train LLMs using target-length texts, but it's relatively rare to find extremely long-context training data. Therefore, how to construct high-quality long context data is important. Another development direction is the advancement of data-level long-context LLMs.

\subsection{Data-level Long-Context LLMs}
Data-level long-context LLMs rely on constructing long-context data. Current data construction methods are costly, often requiring LLMs or human labor. For instance, to tackle the "lost in the middle" \cite{liu2023lost} issue, Prolong \cite{chen2024long} requires OPT-350m \cite{zhang2022opt} to create its large-scale long-context pre-training dataset. LOGO \cite{tang2024logo} requires prompting Qwen2-70B-Instruct \cite{yang2024qwen2} to generate questions for each data instance in its 0.3B token dataset. \cite{xiong2023effective} needs to prompt LLAMA 2 CHAT to generate synthetic self-instruct \cite{wang2022self} long data. LongAlign \cite{bai2024longalign} is constructed via prompting Claude 2.1. 
FILM-7B \cite{an2024make} leverages a long-context QA dataset synthesized by GPT-4-Turbo \cite{achiam2023gpt}. Llama-3-8B-Instruct-QLoRA-80K \cite{zhang2024extending} prompts GPT-4 \cite{achiam2023gpt} to synthesize its 3.5K long-context instruction tuning data. Prolong-8B-Instruct-512k \cite{gao2024train} goes through continual pretraining on 40B token data from Llama3 to obtain long-context abilities. However, these resulting datasets are limited in length and diversity, and the resulting long-context LLMs generally suffer from severe drops in short-context abilities, despite many short contexts having been contained.
 

 Unlike previous data-level methods, Flora is the first to eliminate human and LLM intervention in long-context data construction, enabling theoretically infinite-length contexts with rich diversity while preserving short-context abilities. These unique characteristics of Flora distinguish it from concatenation and data engineering methods, such as LifeLongICL \cite{xu2024stress} and Data Engineering \cite{fu2024data}. LifeLongICL evaluates LLMs' ability to retrieve relevant examples from concatenated few-shot demonstrations for answering new questions. This 'task haystack' approach focuses on assessing task retrieval skills of LLMs. Data Engineering holds that up-sampling long sequences while retaining the domain mixture of the pretraining corpora is crucial for context scaling during pretraining.


\section{Methodology}
\label{headings}

\subsection{Preliminaries}


\textbf{Concatenation Method}: To illustrate data concatenation methods, we reference Mosaic Instruction Tuning (Mosaic-IT) \cite{li2024mosaic}, a technique designed to enhance LLMs' \textbf{instruction-following} capabilities. Mosaic-IT combines multiple short instruction inputs with a meta-instruction to form the input. The output is a meta-instruction-guided concatenation of selected short instruction outputs. 

\textbf{Training Objective}: We first express the objective function of ordinary SFT to lay the groundwork for the introduction of our strategy. Given an SFT dataset $D$ with $n$ data samples, we can divide each sample into a triplet: $(\textit{Instruction}, \textit{Input}, \textit{Response})$. For simplicity, we define $x= (\textit{Instruction}, \textit{Input})$ as the unified instruction, and $y$ as the corresponding \textit{Response}. Let $p_{\theta}(\cdot)$ denote the LLM with parameters $\theta$ that we aim to train. $p_{\theta}$ is fine-tuned by maximizing the objective function $\mathop{\max}\limits_{\theta} \sum_{i=1}^{n} \sum_{j=1}^{l_i} \log p_{\theta}(y_{i,j}|x_i, y_{i,<j})$ over all $N$ samples given as $(x_{i},y_{i})$. Here, $y_{i,j}$ represents the j-th token of the response $y_{i}$, $y_{i,<j}$ denotes the sequence of tokens preceding $y_{i, j}$, and $l_i$ indicates the token length of $y_i$.

\begin{figure*}
    \centering \includegraphics[width=0.96\textwidth,height=0.58\textwidth]{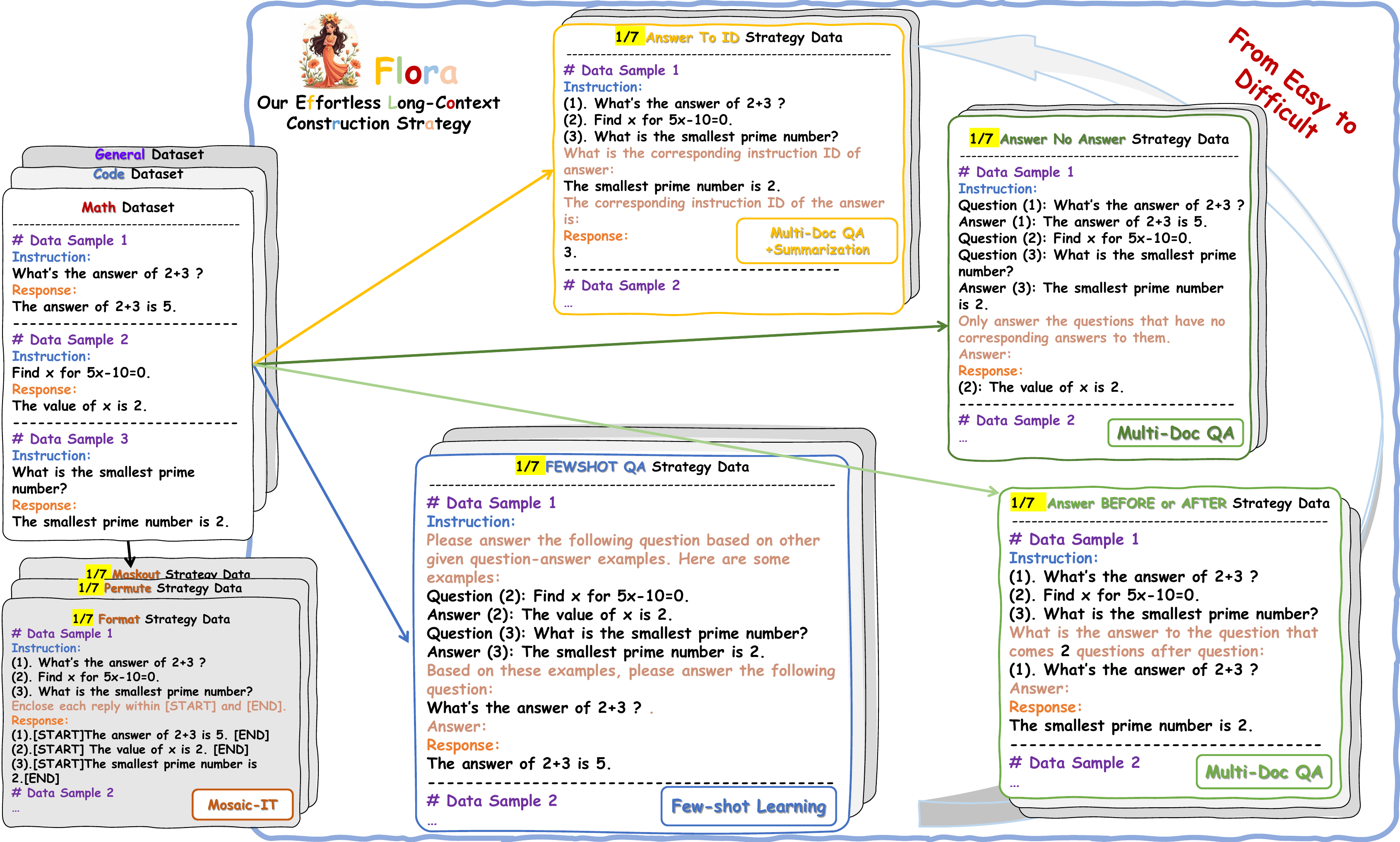}
    \caption{Illustration of Flora, our effortless long-context construction strategy designed to generate theoretically infinite long-context data without any human or LLM intervention. Flora first categorizes the original short instruction tuning datasets into three domains: math, coding, and general knowledge. By applying targeted long-context data augmentations to each category, Flora can enhance the different long-context capabilities of LLMs.}\label{flora}
\end{figure*}

\subsection{Effortless Long-Context Construction} 

Our effortless long-context construction strategy, Flora, incorporates seven data augmentations along with a token length distribution rule, as illustrated in Fig.~\ref{flora}. Each of the seven augmentations contributes 1/7. Three of them are from Mosaic-IT to enhance the instruction following abilities, and the other four are proposed by us to enhance the three most critical long-context capabilities \cite{bai2023longbench} of LLMs: multi-document retrieval, few-shot learning and summarization. Other long-context capabilities can be transferred from the three basic ones. The four long-context strategies, organized from easy to difficult in terms of learning difficulty, include: the Fewshot QA (FQA) strategy to boost few-shot learning skills; the Answer Before or After (ABA) and Answer No Answer (ANA) strategies to improve multi-document retrieval proficiency; and the Answer to ID (AID) strategy to enhance both multi-document retrieval and summarization capabilities. We further explore a token length distribution rule of long-context datasets that can bring better long-context performance gains as a takeaway for constructing long-context datasets. Appendix C provides specific augmented samples and meta-instructions of our four long-context strategies. 


\subsubsection{Fewshot QA (FQA) Strategy}
Few-shot learning ability is crucial for long-context learning since it boosts the LLM's comprehension and reasoning capabilities. Therefore, we design a Fewshot QA (FQA) strategy to enhance LLM's few-shot learning ability.
In the FQA strategy, some arbitrary instructions are selected in the meta-instructions as few-shot examples and require LLMs
to answer new instructions based on given examples. Thus the objective function can be expressed as: $\mathop{max}\limits_{\theta} \sum_{j=1}^{l} \log p_{\theta}([y_{1}^{'},...,y_{\beta}^{'}]_j|[(x_{1},y_{1})...(x_{k},y_{k}),\allowbreak x_{1}^{'},...,x_{\beta}^{'},I_{f}, I_{fqa}], [y_{1}^{'},...,y_{\beta}^{'}]_{< j})$, where $k$ is the count of instructions as few-shot examples that have corresponding responses, $x_{1}^{'},...,x_{\beta}^{'}$ are the $\beta$ instructions we instruct the LLM to answer, $y_{1}^{'},...,y_{\beta}^{'}$ are the corresponding responses, $I_{fqa}$ is the meta-instruction of FQA strategy.

\subsubsection{Answer Before or After (ABA) Strategy}

Designed to improve multi-document retrieval, this approach helps LLMs determine the relevance of information based on its position relative to other content, whether it appears before or after key context. Specifically, we instruct LLMs to answer questions that come $n_{i}$ before or after question $x_{i}^{'}$, where there are $\beta$ questions that need to be answered. Here, $n_{i} \in \{n_{1},...,n_{\beta}\}$, and $x_{i}^{'} \in \{x_{i}^{'},...,x_{\beta}^{'}\}$ is the arbitrarily selected question. For example, this could involve asking the LLMs to answer the question that comes 2 questions after the question "What’s the answer to 2+3?". Thus the objective function can be formulated as: $\mathop{max}\limits_{\theta} \sum_{j=1}^{l} \log p_{\theta}([y_{1}^{'},...,y_{aba}^{'}]_j|[x_{1},...,x_{k},x_{i}^{'},...,\allowbreak x_{\beta}^{'},n_{1},...,n_{\beta},I_{f}, I_{aba}],
[y_{1}^{'},...,y_{\beta}^{'}]_{< j})$, where $I_{aba}$ is the meta-instruction of ABA strategy.

\subsubsection{Answer No Answer (ANA) Strategy}
Also focused on multi-document retrieval, this strategy will arbitrarily concatenate multiple questions and 4/5 of their answers into one instruction and instruct LLMs to only answer the 1/5 questions without corresponding answers. In this way, the LLM can learn to recognize whether the key information is presented in the documents. We define $x_{1}^{'},...,x_{\beta}^{'}$ as the $\beta$ concatenated instructions without corresponding answers $y_{1}^{'},...,y_{\beta}^{'}$. The objective function of ANA can be formulated as: $\mathop{max}\limits_{\theta} \sum_{j=1}^{l} \log p_{\theta}([y_{1}^{'},...,y_{\beta}^{'}]_j|[(x_{1},y_{1})...(x_{k},y_{k}),\allowbreak  x_{1}^{'},...,x_{\beta}^{'},I_{f}, I_{ana}], [y_{1}^{'},...,y_{\beta}^{'}]_{< j})$, where $I_{ana}$ is the meta-instruction of ANA strategy, $k$ is the number of instructions with concatenated answers.

\subsubsection{Answer to ID (AID) Strategy}
This strategy aims to improve multi-document retrieval and summarization abilities. It involves randomly concatenating several questions into a single instruction, excluding their answers. The LLM is then tasked with identifying the question IDs given arbitrary answers to these questions. This approach requires the LLM to comprehend and summarize the answer, as well as retrieve the relevant question from among the multiple concatenated questions. We define $k$ as the number of concatenated questions in an augmented instruction.  $y_{1},...,y_{\beta}$ are the $\beta$ answers given to the LLM to find the corresponding question IDs 
Let $y_{1},...,y_{\beta}$ represent the $\beta$ answers provided to the LLM to determine the corresponding question IDs $y_{1}^{'},...,y_{\beta}^{'}$. Thus the objective function of AID can be formulated as: $\mathop{max}\limits_{\theta} \sum_{j=1}^{l} \log p_{\theta}([y_{1}^{'},...,y_{\beta}^{'}]_j|[x_{1},...x_{k}],[y_{1},...y_{\beta}\allowbreak , I_{f}, I_{aid}],
[y_{1}^{'},...,y_{\beta}^{'}]_{< j})$, where $I_{aid}$ is the meta-instruction of AID strategy.

\subsubsection{Long-Context Dataset Construction}

\begin{figure}
    \centering \includegraphics[width=0.48\textwidth,height=0.20\textwidth]{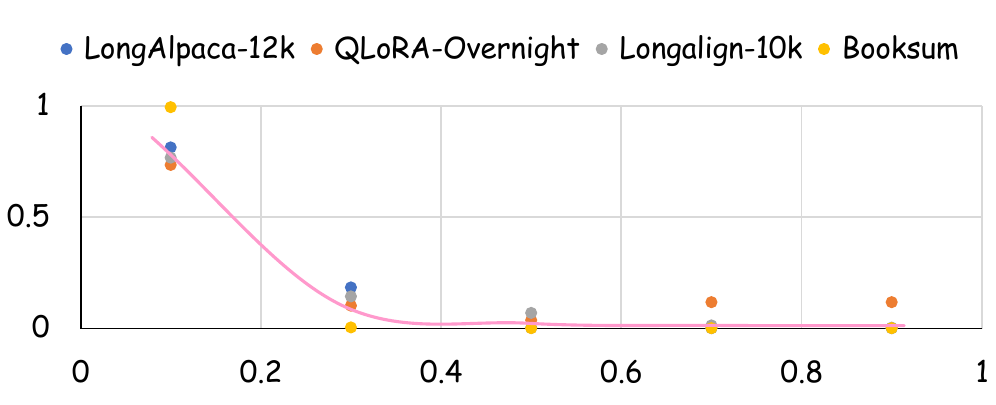}
    \caption{We show that the token length distribution of present public long context instruction tuning datasets can be fitted as a curve: $y = 2.411e^{-10.899x}+0.017$, where $x$-axis measures the normalized token length ranges from 0 to 1, and $y$-axis measures the data sample proportion. }\label{function}
\end{figure}

We analyze some most widely used long-context SFT datasets (LongAlpaca-12k \cite{chen2023longlora}, LongAlign-10k \cite{bai2024longalign}, Booksum \cite{kryscinski2021booksum} and QLoRA-Overnight \cite{zhang2024extending}) and find that their ideal token length distribution follows $y = 2.411e^{-10.899x}+0.017$, as shown in Fig.~\ref{function}, where $x$ is the normalized token length (0 to 1) and $y$ is the data sample proportion. We follow this ideal distribution to construct our dataset. 


We adopt the Infinity Instruct\footnote{https://huggingface.co/datasets/BAAI/Infinity-Instruct} dataset, which has around 1.5 million entries composed of extensive open-source short-length SFT data samples collected by BAAI and enhance it with our Flora strategy to be the final long-context datasets. The token length distribution adheres to our functional rule. We categorize the Infinity Instruct dataset into math, code, and general knowledge. Each category is then enhanced with QAF, ABA, ANA, and AID augmentations, plus 3 Mosaic augmentations, before merging the augmented categories. To better preserve short-context capabilities, we also incorporate original short data samples (those under 2k tokens) from the Infinity Instruct to replace the augmented samples under 2k tokens. 

We compare our datasets with some other widely used public long-context datasets in Table~\ref{dataset_compare}, where our curated datasets cover more domains and is flexible in scale and length. We have omitted the thinking mode token during curating the dataset for QwQ, therefore training on QwQ can be regarded as an instruction tuning process. The model after SFT can still generate the thinking part.

\begin{table}
\begin{center}
\centering
\resizebox{0.99\columnwidth}{!}{
    \begin{tabular}{c|cccc}
    \toprule
        Datasets & Domain & Sample Num & Max Token & Avg Token\\
        \midrule
LongAlpaca-12k& 1 &12k & 70k & 9.4k\\
LongAlign-10k& 6&10k & 86k & 16.9k\\
QLoRA-Overnight& 7&20k & 80k & 14k\\
Flora-80k (Ours)&15 &93k & 80k &9.5k \\
Flora-128k (Ours)&15 &60k & 128k &14.8k \\
        \bottomrule
    \end{tabular}}\label{setting}
\end{center}
\caption{Comparison with other widely used long-context datasets in terms of domain (e.g., math, code, science, etc.), sample number, and token lengths. The token length is calculated by Llama-3 tokenizer.} \label{dataset_compare}
\end{table}


\begin{table*}

\centering
\resizebox{2.0\columnwidth}{!}{
    \begin{tabular}{c||c>{\columncolor{gray!25}}c|c>{\columncolor{gray!25}}c|c>{\columncolor{gray!25}}c|c>{\columncolor{gray!25}}c|c>{\columncolor{gray!25}}c|cc||ccccc|c}
    \toprule[1pt]
        \multirow{2}{*}{\textbf{Models}} &\multicolumn{4}{c|}{\textbf{Difficulty}} & \multicolumn{6}{c|}{\textbf{Length (<32k; 32k-128k; >128k words)}} &\multicolumn{2}{c||}{\multirow{2}{*}{\textbf{\shortstack{LongBench \\ v2 Avg.}}}}& \multirow{2}{*}{\textbf{SQA} }& \multirow{2}{*}{\textbf{MQA}} & \multirow{2}{*}{\textbf{Summ}}& 
        \multirow{2}{*}{\textbf{FS}}& 
        \multirow{2}{*}{\textbf{Syn}}& \multirow{2}{*}{\textbf{\shortstack{LongBench \\ v1 Avg.}}}\\
        &\multicolumn{2}{c}{Easy} & \multicolumn{2}{c|}{Hard}& \multicolumn{2}{c}{Short}& \multicolumn{2}{c}{Medium}& \multicolumn{2}{c|}{Long}& & & & & & &\\
        \midrule
        \rowcolor{blue!10}
        \multicolumn{19}{c}{\textit{Model-Level Comparison with SOTA Long-Context LLMs}}\\
        
        GLM-4-9B-Chat-128k&30.7 &34.4& 29.9& 28.6& 33.9& 35.0 &29.8 &30.2 &25.0 &25.0& 30.2&\cellcolor{gray!25}30.4 &47.30 &42.7 &23.94 &40.61 &99&50.71\\
        Llama-3.1-8B-Instruct-128k &30.7& 36.5& 29.6& 26.7& 35.0 &34.4& 27.9& 31.6& 25.9& 21.3& 30.0 & \cellcolor{gray!25}30.4 & 47.71 &37.46& 23.75 &42.68&98.5&50.02\\
        Qwen2.5-7B-Instruct&29.2& 30.7& 25.7& 29.3& 36.1 &35.6 &23.7 &26.5 &18.5 &26.9&27.0 & \cellcolor{gray!25}29.8 & 41.00 &37.74& 22.54 &43.2 &84.84 &45.86\\
        Llama-3-8B-Instruct-262k &35.9		&32.8 & 28.3&26.4 & 33.3 &35.6  &31.6 &25.6 &26.9 & 24.1&31.2  & \cellcolor{gray!25}28.8 &35.69  & 18.02& 17.10 & 40.59 &88.5 & 39.98\\
        Claude-3.5-Sonnet-200k &46.9	&55.2	&37.3	&41.5	&46.1	&53.9	&38.6	&41.9	&37.0	&44.4 &41.0	&\cellcolor{gray!25}46.7& -&- &- &- &- &- \\
        Qwen3-235B-A22B-128k $\clubsuit$&47.4	&56.4	&36.0	&46.2	&45.6	&58.3	&36.7	&44.1	&38.9	&48.6 &40.4&\cellcolor{gray!25}50.1 & -&- &- &- &- &- \\		
        \rowcolor{yellow!20}
        \multicolumn{19}{c}{\textit{Data-Level Comparison with Long-Context Datasets}}\\   
        InternLM2-7B-LongWanjuan* & - &-&-&-&-&-& - &-&-&-&-&\cellcolor{gray!25}-& 46.22& 37.22& 27.63& 41.02& 93&49.01 \\
        Mistral-FILM-7B & 25.7 &27.7 &18.3 &21.5 &23.9 &24.6 & 20.5 & 25.0 &17.8 &20.4 &21.1&\cellcolor{gray!25}23.9  & 48.46 & 38.95& 24& 39.70& 95&49.32\\
        Llama-3.1-8B-Instruct-SEALONG & 29.7	 &37.0&26.4&27.3&32.8&36.7& 24.2 &28.8&25.9&25.9&27.6&\cellcolor{gray!25}31.0 & 48.25& 37.83 & 24.38 & 42.5 & 98.5  & 50.29\\
Prolong-8B-Instruct-512k (CPT) & 30.9 &31.6 &25.2&22.2 &32.4 &28.7 & 27.5 &25.5 &18.6&21.4 &27.3&\cellcolor{gray!25}25.8
 & 44.44& 20.03 & 24.76 & 43.24 & 95.75& 45.64\\
Llama-3-8B-Instruct-QLoRA-80k & 32.3& 27.6& 25.1& 20.6& 30.0& 27.2& 29.8& 22.3& 20.4& 18.5& 27.8& \cellcolor{gray!25}23.3& 45.49& 35.33& 15.40& 40.92& 94.5&46.33 \\
Llama-3-8B-Instruct-LifeLongICL &29.9 &25.9 &26.6 &29.4 &34.9 &36.7 &25.4 &23.6 &20.6 &22.1 &27.8 & \cellcolor{gray!25}28.1 & 37.28&23.04 &13.68 &38.4 & 98&42.08 \\

\midrule
\textit{\textbf{Llama-3-8B-Instruct-8k }} & \textit{0} &\textit{0}&\textit{0}&\textit{0}&\textit{0}&\textit{0}& \textit{0} &\textit{0}&\textit{0}&\textit{0}&\textit{0}&\cellcolor{gray!25}\textit{0} &\textit{40.59} & \textit{28.34} & \textit{14.05} & \textit{32.67} & \textit{79}  & \textit{38.93}\\
+ LongAlpaca-12k &29.1 &29.2 &28.7 & 25.1 &36.1 & 32.8&26.1 &23.3 &22.2 &23.1 &28.9 & \cellcolor{gray!25}26.6 &44.58 &34.98 & 22.7&41.30 & 94.25& 47.56\\
+ LongAlign-10k &29.5 &28.3 &29.1 &30.7 & 32.7&32.5 &26.7 &27.7 &28.7 & 29.3& 29.3& \cellcolor{gray!25}29.8&  44.38&27.36 &22.5 &39.65 &77.13 &42.24 \\

+ Flora-80k (Ours) & 33.3& 34.9& 33.4& 32.2& 45.0& 35.0& 30.7& 33.0& 19.4 & 30.6&33.4 & \cellcolor{gray!25}33.2& 48.68 &39.36&23.6&42.51&99.5&50.73\\
\midrule
    \textit{\textbf{QwQ-32B-128k} $\clubsuit$} & - & \textit{50.6}& -&\textit{42.6} &- &\textit{48.8}  &- &\textit{45.2} &- &\textit{40.8} &- &\cellcolor{gray!25}\textit{45.6} & \textit{83.11}&\textit{77.54} &\textit{26.78} &\textit{44.95} &\textit{98.75} &\textit{66.23} \\	
+ LongAlpaca-12k $\clubsuit$&- &43.5 &- &42.0  &- &51.5 &- &38.2 &- &36.1 &- & \cellcolor{gray!25}42.5 & 80.05&74.26 &24.82 &41.66 &95.5 &63.26\\	

+ LongAlign-10k $\clubsuit$&- &49.1 &- &43.8  &- &54.5 &- &42.6 &- &37.8 &- & \cellcolor{gray!25}45.8 & 83.21&78.56 &25.87 &42.15 &98 &65.56 \\
+ Flora-128k (Ours) $\clubsuit$&-&\textbf{61.8}  &- &\textbf{44.8}  &- &\textbf{58.8} &- &\textbf{46.2} &- &\textbf{46.4} & -& \cellcolor{gray!25}\textbf{50.5} & \textbf{87.12}&\textbf{83.06} &\textbf{29.38} &\textbf{51.39} &\textbf{99.5} &\textbf{70.07} \\	

        \bottomrule[1pt]
    \end{tabular}
    }
\caption{Results ($\%$) on LongBench v2 and v1:  CoT prompting results in LongBench v2 are highlighted with a \colorbox{gray!25}{gray} background, with random guessing at 25$\%$. Bold numbers indicate the highest values per column. InternLM2-7B-LongWanjuan \cite{lv2024longwanjuan}, marked with *, is closed-source, and only its official LongBench v1 results are reported. Llama-3-8B-Instruct-LifeLongICL denotes the model trained by Flora-enhanced LifeLongICL data. Reasoning models are indicated by $\clubsuit$. QwQ-32B-128k results are self-evaluated and limited to its reasoning mode.} \label{table1}
\end{table*}

\section{Experimental Setup}


\subsection{SFT Details}
We use QLoRA \cite{dettmers2024qlora} to efficiently fine-tune Llama-3-8B-Instruct as our main model based on Llama-Factory\footnote{https://github.com/hiyouga/LLaMA-Factory}. We apply LoRA on all Q, K, V, and O projections
and additionally train the embedding layer. The LoRA rank is set to 256, with an alpha value of 128 and 4-bit quantization. The learning rate is set to 5e-5, with a linear decay schedule and no warm-up steps. The batch size is 8, and gradient checkpointing is enabled to optimize memory usage. The model is trained for one epoch with 8.5$\%$ trainable parameters for Llama3-8B and with 5.5$\%$ trainable parameters for QwQ-32B on $4\times 8A100 (80G)$ machines using DeepSpeed v2 offloading within a day. For Llama3-8B, we expand the RoPE base to 200M and increase max position embeddings to 81,920. For QwQ-32B, we keep the original RoPE base and increase max position embeddings to 131,072. The $\beta$ is set to 1 for FQA, ABA and AID, and set to 20$\%$ of the concatenated instructions for ANA.

\subsection{Evaluation Details}
\subsubsection{Compared Methods}
To prove the effectiveness of our strategy, we compare our dataset with other long-context datasets with similar training settings. To showcase the long-context capabilities of our final model, we evaluate it against other prevalent long-context LLMs with comparable parameters. Specifically, we fine-tune Llama-3-8B-Instruct and QwQ-32B \cite{team2025qwq}\footnote{For the result mismatch of our self-evaluated QwQ-32B and the LongBench v2 leaderboard result, we strictly followed the author’s official hyper-parameter recommendations in a Github issue: Temperature=0.6, TopP=0.95, MinP=0, max new tokens=30000, max input len=100000, timeout=3600, and enabling YaRN whilst evaluating.} on long-text datasets: LongAlpaca-12k \cite{chen2023longlora}, LongAlign-10k \cite{bai2024longalign} and our Flora dataset with maximum token lengths of 80k and 128k. This demonstrates the scalability of our strategy across model parameters and token lengths. Other data-level LLMs in comparison include Llama-3-8B-Instruct-QLoRA-80k \cite{zhang2024extending}, Prolong-8B-Instruct-512k \cite{gao2024train}, InternLM2-7B-LongWanjuan \cite{lv2024longwanjuan}, Mistral-FILM-7B \cite{an2024make}, Llama-3.1-8B-Instruct-SEALONG \cite{li2024large} and our self-trained Llama-3-8B-Instruct on the data of LifeLongICL \cite{xu2024stress} enhanced by our Flora. For model comparison, we compare our model with other long-context LLMs trained to handle context windows $\geq$ 32k tokens (GLM-4-9B-Chat-128k \cite{glm2024chatglm}, Qwen-2.5-7B-Instruct \cite{qwen2.5}, Llama-3.1-8B-Instruct-128k \cite{dubey2024llama}, ChatGLM3-6B-32k \cite{glm2024chatglm}, 
Llama-3-8B-Instruct-262k \footnote{https://huggingface.co/gradientai/Llama-3-8B-Instruct-262k}, Claude-3.5-Sonnet, and Qwen3-235B-A22B-128k \footnote{https://github.com/QwenLM/Qwen3}.

\begin{figure*}
    \centering \includegraphics[width=0.99\textwidth,height=0.36\textwidth]{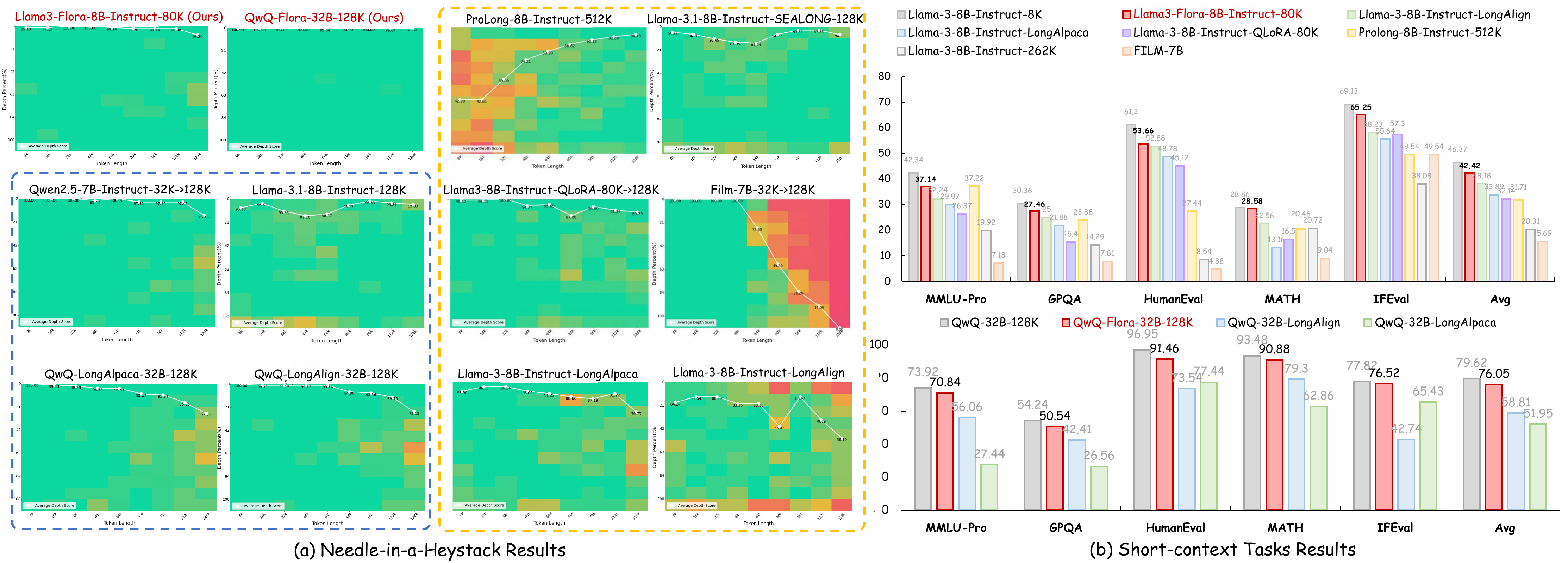}
    \caption{(a) Results of Single-Retrieval task in Needle-In-A-Haystack benchmark. The $x$-axis represents the context lengths, while the $y$-axis indicates the depth of the inserted needle. The green color signifies a score close to 1, and red denotes a score close to 0. (b) Results on five short-context tasks from the Open LLM Leaderboard 2.}\label{needle}
\end{figure*}

\subsubsection{Long-context Benchmarks}
We adopt three benchmarks, LongBench v1 \cite{bai2023longbench}, LongBench v2 \cite{bai2024longbench} and Needle-In-A-HayStack \footnote{https://github.com/gkamradt/LLMTest$\_$NeedleInAHaystack}, to evaluate the long-context understanding ability of various LLMs. 

LongBench v1 is widely used to evaluate long-context LLMs in handling inputs mostly below 20k words. We take 11 representative tasks from it to form 5 task types, including single-document
question answering (SQA), multi-document question answering (MQA), summarization (Summ), few-shot learning (FS) and synthetic tasks (Syn). The selected tasks are given in Appendix A. 

However, significant advancements in long-context LLMs have increased  context window lengths significantly from 8k to 128k and even up to 1M tokens, making LongBench v1 inadequate to evaluate LLMs capable of handling more than 20k words. To address this, LongBench v2 and Needle-In-A-HayStack benchmarks are adopted. LongBench v2 includes 503 challenging multiple-choice questions across contexts ranging from 8k to 2 million words, with most below 128k words. It focuses on deep understanding and reasoning across real-world multitasks and offers a more comprehensive and challenging assessment than v1. It can also serve as an OOD benchmark, as it is designed to assess (1) long-context reasoning and (2) the understanding of long structured data (including reasoning over lengthy tables and knowledge graphs)-task categories that are not directly emphasized by our concatenation strategy. Needle-In-A-HayStack challenges LLMs to recall irrelevant information inserted into a lengthy context and is assessed by GPT-3.5. We test the single needle retrieval tasks. For tested models with original maximum position embeddings below 128k, we extend their embeddings to 128k for these evaluations.

\subsubsection{Short-context Benchmarks}
We select 5 tasks from Open LLM Leaderboard v2 to evaluate the short-context abilities of models, including \textbf{MMLU-Pro} \cite{wang2024mmlu}, \textbf{GPQA} \cite{rein2023gpqa},\textbf{IFEval} \cite{zhou2023instruction}, \textbf{HumanEval} \cite{chen2021evaluating} and \textbf{MATH} \cite{hendrycks2021measuring}. The first three are employed to test the general knowledge abilities, while \textbf{MATH} \cite{hendrycks2021measuring} and \textbf{HumanEval} \cite{chen2021evaluating} are used to test math and coding abilities, respectively. We employ the OpenCompass library \footnote{https://github.com/open-compass/opencompass} to evaluate all these tasks.

\section{Experiments and Analysis}
\subsection{Long Context Results}
We evaluate our Flora-enhanced LLMs against a series of the latest long-context LLMs. This comparison encompasses both prevalent LLMs and data-level LLMs to which our model belongs. The results on LongBench v2 and v1 are detailed in Table~\ref{table1}, where our model achieves SOTA performances on average on both benchmarks. Since the overall test length of LongBench v1 mainly covers the short length parts of LongBench v2, LongBench v2 can better reflect the long-context understanding and reasoning abilities of different LLMs. Notably, Llama-3-8B-Instruct-8k struggles with these demanding long-context challenges and outputs null results, resulting in zero scores on LongBench v2. Compared to other data-level models, Llama3-Flora-8B-Instruct-80k demonstrates significant superiority in handling hard questions and those below 128k words without CoT prompts. With the use of CoT, our model continues to markedly outperform others, even showing an enhanced ability to comprehend and reason through particularly long contexts (beyond 128k words).

The results on Needle-In-A-HayStack are visualized in Fig.~\ref{needle} (a), where our models nearly achieves a 100$\%$ retrieval accuracy across all context lengths and the Llama3 model shows excellent generalization to new positions (80k-128k).

\begin{table*}
\centering
\resizebox{2.0\columnwidth}{!}{
    \begin{tabular}{c||cc>{\columncolor{gray!25}}c|cc>{\columncolor{gray!25}}c|cc>{\columncolor{gray!25}}c|cc>{\columncolor{gray!25}}c|cc>{\columncolor{gray!25}}c|c||c}
    \toprule[1pt]
        \multirow{2}{*}{\textbf{Augmentations (Normal Dist)}} &\multicolumn{3}{c|}{\textbf{SQA}} & \multicolumn{3}{c|}{\textbf{MQA}} & \multicolumn{3}{c|}{\textbf{Summ}}& 
        \multicolumn{3}{c|}{\textbf{FS}}& 
        \multicolumn{3}{c|}{\textbf{Syn}}& \multirow{2}{*}{\textbf{\shortstack{LongBench \\ v1 Avg.}}}&\multirow{2}{*}{\textbf{\shortstack{Short Context \\ Tasks Avg.}}}\\
        &Narr QA &  MQA-zh & \textbf{Avg.} &2Wiki & MSQ  & \textbf{Avg.} & Gov &VCSUM  & \textbf{Avg.} &SAMSum &LSHT  & \textbf{Avg.} &PR-en &PR-zh & \textbf{Avg.} &\\
        \midrule
        Baseline (Mosaic-IT) &25.71  &55.28 & 40.50&45.03 & 27.83& 36.43&23.98 &13.19 & 18.58&37.42 &39.5 & 38.46&94.5 & 93& 93.75&45.54 &42.76\\
        3/4 Baseline + 1/4 \color{blue!50}{FQA} &27.09 &55.16 & 41.13&45.81	 &31.05 & 38.43&25.88 &12.22& 19.05 &\textbf{41.80} &42 & \textbf{41.9}& 96& 97& 96.5&47.40 &42.89\\
        3/4 Baseline + 1/4 ABA &26.4  &55.04 & 40.72&49.73	 &\textbf{33.56} & \textbf{41.65}& 25.26	&13.87&  19.57&39.97 &40 & 39.98&97.5 & 98& 97.75&47.93 &43.03\\
        3/4 Baseline + 1/4 ANA &27 &55.23& 41.12 &49.48 &32.88 & 41.18&26.04 &14.38 & 20.21&38.14	 &41.5 & 39.82& 98.5& 97& 97.75&47.97 &42.87\\
        3/4 Baseline + 1/4 ATID &\textbf{28.41} &53.05 & 40.73&49.15	 &31.67 & 40.41&\textbf{27.85} &15.74 & 21.80&39.43	 &41.5 & 40.46&99 &98 & 98.5&48.38 &42.94\\
        \color{blue!50}{FQA} +ABA +ANA + ATID & 27.52& 54.87& 41.20&48.63 & 29.8& 39.22& 27.09&15.99 & 21.54&39.58 & 41& 40.29&99& \textbf{98.5} & 98.75&48.20 &42.91\\
        3/7 Baseline + 4/7 All &28.10 &55.88 & 41.99&49.51 & 32.06& 40.79& 28.01&16.18 &22.10 &40.66 & 42&41.33 &99&98.5 & 98.75&48.99 &43.02\\ 
        3/7 Baseline + 4/7 All $\dag$ (Ours)&28.24 &\textbf{55.93} & \textbf{42.08}& \textbf{49.76}&31.91 &40.84 &27.82 &\textbf{16.37} & \textbf{22.10}&41.29 &	\textbf{42} & 41.65& \textbf{99.5}& 98& \textbf{98.75}& \textbf{49.08} &\textbf{43.41}\\
        \bottomrule[1pt]
    \end{tabular}
    }
\caption{Ablation studies on different long-context augmentations on LongBench v1 and short context tasks. All the augmented datasets contain 50k data samples, and the token lengths follow a normal distribution for fair comparisons. $\dag$ means the augmented samples below 2k tokens are replaced by the original SFT samples below 2k tokens.} \label{ab1}
\end{table*}

\begin{table*}
\centering
\resizebox{2.0\columnwidth}{!}{
    \begin{tabular}{c||cc|ccc|c||ccccc|c||c}
    \toprule[1pt]
        \multirow{2}{*}{\textbf{Model}} &\multicolumn{2}{c|}{\textbf{Difficulty}} & \multicolumn{3}{c|}{\textbf{Length }} &\multirow{2}{*}{{\textbf{\shortstack{LongBench \\ v2 Avg.}}}}& \multirow{2}{*}{\textbf{SQA} }& \multirow{2}{*}{\textbf{MQA}} & \multirow{2}{*}{\textbf{Summ}}& 
        \multirow{2}{*}{\textbf{FS}}& 
        \multirow{2}{*}{\textbf{Syn}}& \multirow{2}{*}{{\textbf{\shortstack{LongBench \\ v1 Avg.}}}} &\multirow{2}{*}{\textbf{\shortstack{Short Context \\ Tasks Avg.}}}\\
        & Easy & Hard& Short& Medium& Long& & & & & & & &\\
        \midrule
        Baseline (Llama-3-8B-Instruct-8k)
        &  0& 0& 0& 0& 0& 0&40.59 &28.34 &14.05 &32.67& 79& 38.93 &46.37\\  
        Normal Distribution &\textbf{33.5}	&28.7	&36.6	&27.7	&25.8 &30.5& 44.77& 38.01& 21.14 & 39.86 & 98.25 & 48.40 & 43.56\\
        
        $y = 0.2$ (Even) &30.7	&29.9	&\textbf{37.8}	&29.8	&18.5 &30.2 & 44.54&38.80 & 21.01& 39.94& 97&48.26 &   43.83 \\
        
        $y = 2.375(x-0.5)^2+0.01$ (U-Shaped) &32.8	&28.0	&36.7	&28.4	&21.3 &29.8 &41.73 &\textbf{40.33} & 20.03& 38.84& 98.5 &47.89  &44.18 \\

        $y = 2.411e^{10.899(x-1)}+0.017$ (Reverse) & 31.8 &26.9 &33.3 &28.2 &22.2 &28.7 & 43.90 &38.30 &20.24 & 40.13 &97.25 & 47.96 &40.65 \\
        
        $y = 2.411e^{-10.899x}+0.017$ (Ours) &33.0	&\textbf{31.8}	&37.4	&\textbf{30.2}	&\textbf{27.8}&\textbf{32.3} &\textbf{45.05}  &38.69 &\textbf{21.69} &\textbf{41.95 }&\textbf{98.75}& \textbf{49.23} &\textbf{44.82} \\   
        \bottomrule[1pt]
    \end{tabular}
    }%
\caption{Ablation studies on different token length distribution on LongBench v2, v1 and short context tasks. All experiments use 8k samples, the same number as the dataset of the reverse distribution, for fair comparison.} \label{ab2}
\end{table*}

\subsection{Short Context Results}

Besides long-context assessments, we also test several long-context LLMs on five short-context tasks from the Open LLM Leaderboard 2 in Fig.~\ref{needle} (b). All Llama3 and QwQ series models perform worse than baselines, suggesting that extending the context window compromises the model's ability to handle short contexts. This observation aligns with previous researches \cite{peng2023yarn,zhang2024extending}. Notably, the performance decline of other methods on short-context tasks is not due to a lack of short-context data. For instance, LongAlpaca-12k includes a significant amount of such data yet still underperforms. For LongAlign, its training setting is different from ours: it first conducts continual pretraining over 10B tokens (with sequence lengths up to 64k), followed by instruction tuning on its long-context dataset LongAlign-10k with long and short context data. As our strategy focuses on the SFT stage, we ensure a fair comparison by evaluating directly against its SFT dataset LongAlign-10k.

However, our model boasts a significantly smaller performance decline compared to other similarly scaled long-context LLMs, with only a 3-4$\%$ average drop versus at least a 8$\%$ average drop for other models. It excels in retaining basic knowledge, coding, math, and instruction-following abilities. This is because our strategy involves splitting the original instruction tuning dataset into three categories before concatenating samples with diverse meta-instructions, while keeping samples under 2k tokens as original instruction tuning dataset samples. By concatenating samples of the same category, we preserve domain-specific knowledge. The arbitrary sample concatenation, along with the use of varied meta-instructions, helps maintain instruction-following skills. Retaining shorter samples as part of the original data also contributes to this preservation.

\subsection{Ablation Studies}

Table~\ref{ab1} presents an ablation study on various long-context augmentation strategies, showcasing the average results across different evaluation criteria on LongBench v1 and short-context tasks. Note that we only selected LongBench v1 for our experiments because it includes tasks designed to assess the corresponding effect of our augmentation strategies. We adopt Mosaic-IT as our baseline and concatenate the maximum data length to 80k tokens for fair comparison. The dataset sample length adheres to a normal distribution as Mosaic-IT and all the augmented datasets contain 50k data samples in Table~\ref{ab1}. To evaluate the effectiveness of each augmentation strategy, we incorporate each strategy once, allowing its augmented data to constitute 1/4 of the total samples, while the remaining 3/4 consists of data augmented by Mosaic-IT. The results clearly demonstrate that each of our proposed strategies significantly enhances the corresponding long-context abilities of the original LLM. Implementing all four strategies together leverages their individual strengths, resulting in the best overall performance. Replacing the very short concatenated contexts with the original samples can better maintain short-context abilities.

We also investigate how different token length distributions affect LLM's understanding of long and short contexts. The findings, detailed in Table~\ref{ab2}, are based on experiments using datasets of 8k samples, each with a maximum token length of 80k, ensuring a fair comparison. We analyze various distributions: normal, even, U-shaped, reverse, and our optimized rule. 
The results suggest that more short samples mitigates the decline in short-context understanding, while only a few extremely long samples are needed for excellent long-context capabilities. Our approach yields optimal proficiency in both long and short contexts.

    



\section{Conclusion}
Handling extremely long contexts remains a challenge for LLMs due to the scarcity of such data, high computational demands, and the issue of catastrophic forgetting of short contexts. Existing methods often involve costly and limited human or model intervention. We introduce Flora, a new approach for constructing long contexts without human or model involvement. Flora assembles short instruction contexts into theoretically infinite-length contexts, paired with high-level meta-instructions for training. This method enhances long-context capabilities with minimal impact on short-context abilities. 

\section{Limitations}


Our research introduces an innovative, human/LLM-independent strategy for long-context data construction. This approach can generate theoretically unlimited context lengths and significantly enhance long-context capabilities while minimally impacting short-context performance. However, its current application is primarily confined to the language domain. Future research should explore extending this strategy to multi-modal and visual fields, broadening its applicability across diverse domains. We have defined four meta-instruction types within our strategy, laying a foundation for future expansion in the future to improve diversity and generalizability. Due to computational constraints, we have yet to investigate the scalability of our approach to larger language models (70B parameters or more) and more extensive datasets.




\bibliography{main}
\newpage
\section*{Appendices}
\appendix
\section{11 Tasks from LongBench v1}
\label{sec:appendix}
We take 11 representative tasks to form 5 task categories from LongBench v1, including single-document
question answering (Narrative QA \cite{kovcisky2018narrativeqa}, Multi-FieldQA-en, Multi-FieldQA-zh \cite{bai2023longbench}, multi-document question answering (2WikiMultihopQA \cite{ho2020constructing} and MuSiQue \cite{trivedi2022musique}), 
long-context summarization (GovReport \cite{huang2021efficient}, QMSum \cite{zhong2021qmsum}, few-shot learning (SAMSum \cite{gliwa2019samsum}, lsht\cite{bai2023longbench}) and synthetic tasks (PassageRetrieval-en, PassageRetrievalzh \cite{bai2023longbench}).

\section{Infinity Instruct Dataset}
BAAI gathered extensive open-source data as seed material, refining it through instruction selection and evolution. We utilize the InfInstruct-Gen (0729) dataset, evolved from a high-quality seed subset with approximately 1.5 million entries, to enhance the model's instruction-following ability in real-world conversational scenarios.

\section{Token Length Distribution Details}
For a dataset with a maximum token length of 80k, according to our proposed token length distribution rule, samples are distributed as follows: 0-16k tokens account for approximately 82.8$\%$, 16-32k tokens for 10.9$\%$, 32-48k tokens for 2.8$\%$, 48-64k tokens for 1.8$\%$, and 64-80k tokens for 1.7$\%$. In the ablation study part, samples of a U-shaped distribution are distributed as follows: token lengths of 0-16k make up approximately 39$\%$, 16-32k for 10.5$\%$, 32-48k for 1$\%$, 48-64k for 10.5$\%$, and 64-80k for 39$\%$. The reverse distribution allocates around 1.7$\%$ for 0-16k tokens, 1.8$\%$ for 16-32k, 2.8$\%$ for 32-48k, 10.9$\%$ for 48-64k, and 82.8$\%$ for 64-80k. 

\section{Domain Coverage of different datasets}
We prompt GPT-4o to classify the domain of present long-context SFT datasets and our curated datasets. The covered domain of QLoRA-Overnight includes code, math, literature, language, tech, history, philosophy. The covered domain of Longalpaca-12k includes mainly scientific literature. The covered domains of LongAlign-10k include history, politics, code, math, literature, news, tech, finance, law. The covered domains of our datasets includes history, politics, code, math, literature, news, tech, finance, law, social comments, philosophy, scientific papers, health, art, religion



\section{Specific Examples}

\begin{figure}[H]
    \centering \includegraphics[width=0.48\textwidth,height=0.63\textwidth]{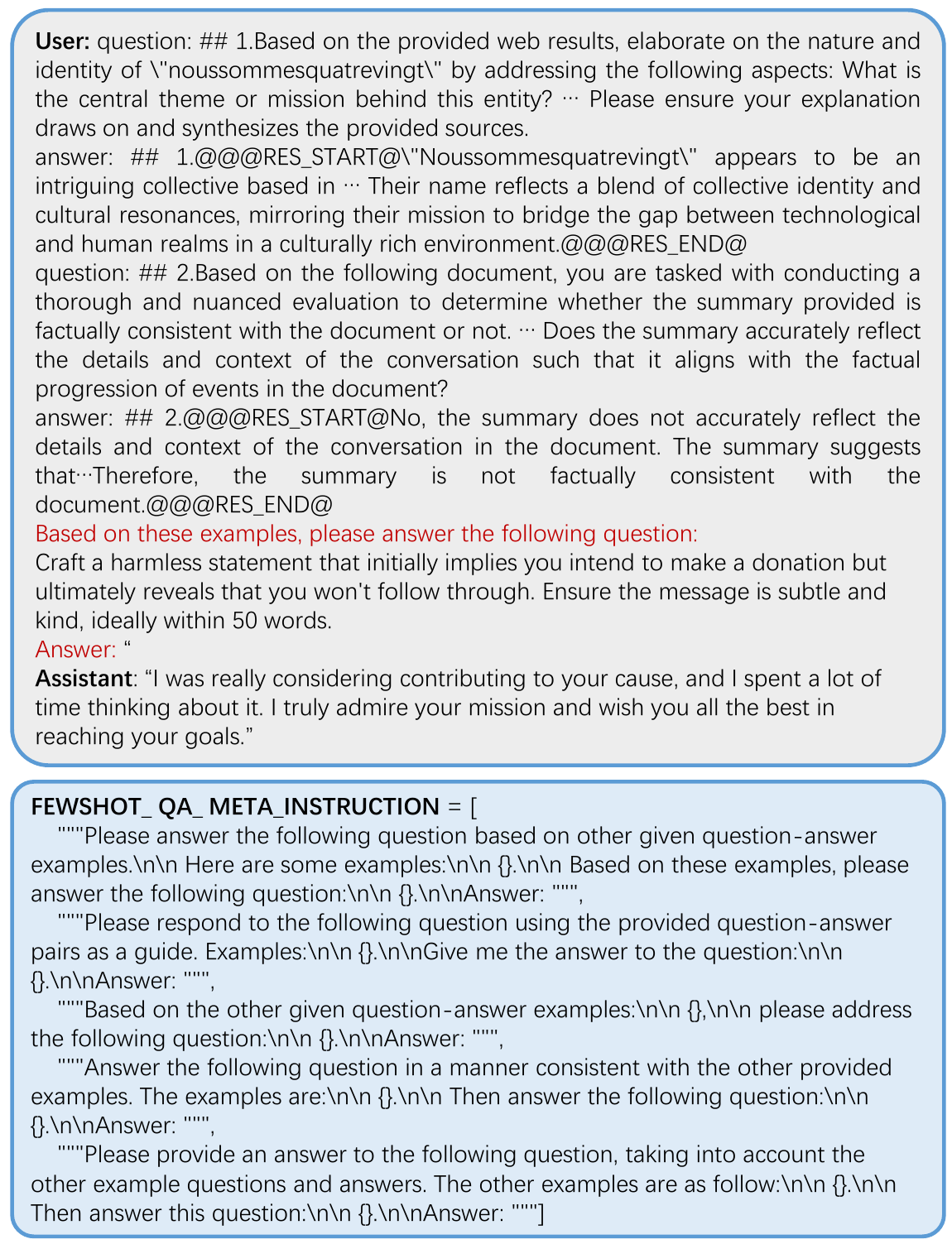}
    \caption{An Example of the FQA augmented data sample and meta-instructions.}\label{example_fewshot}
\end{figure}

\begin{figure}
    \centering \includegraphics[width=0.5\textwidth,height=0.40\textwidth]{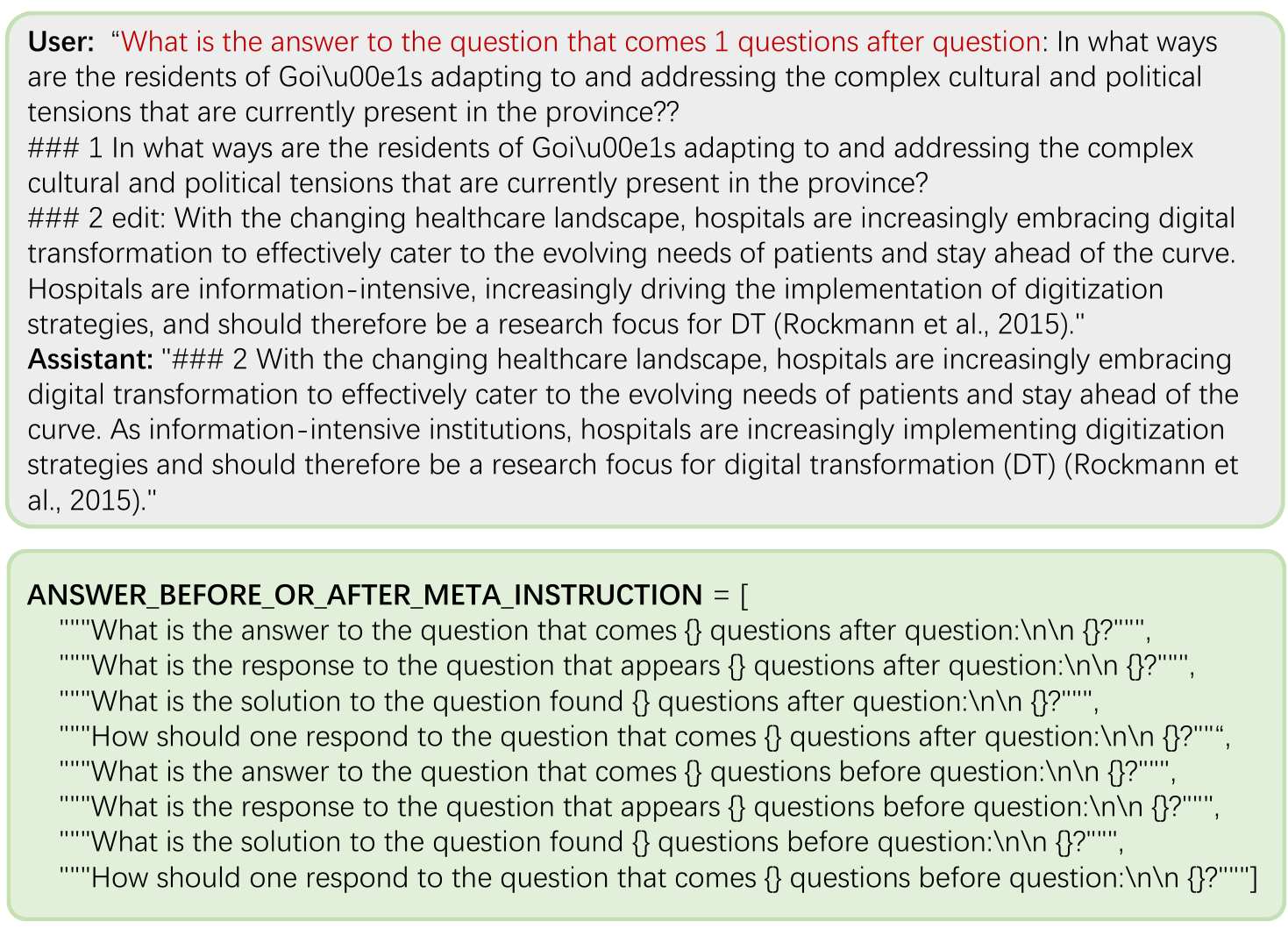}
    \caption{An Example of the ABA augmented data sample and meta-instructions.}\label{example_aba}
\end{figure}

\begin{figure}
    \centering \includegraphics[width=0.48\textwidth,height=0.67\textwidth]{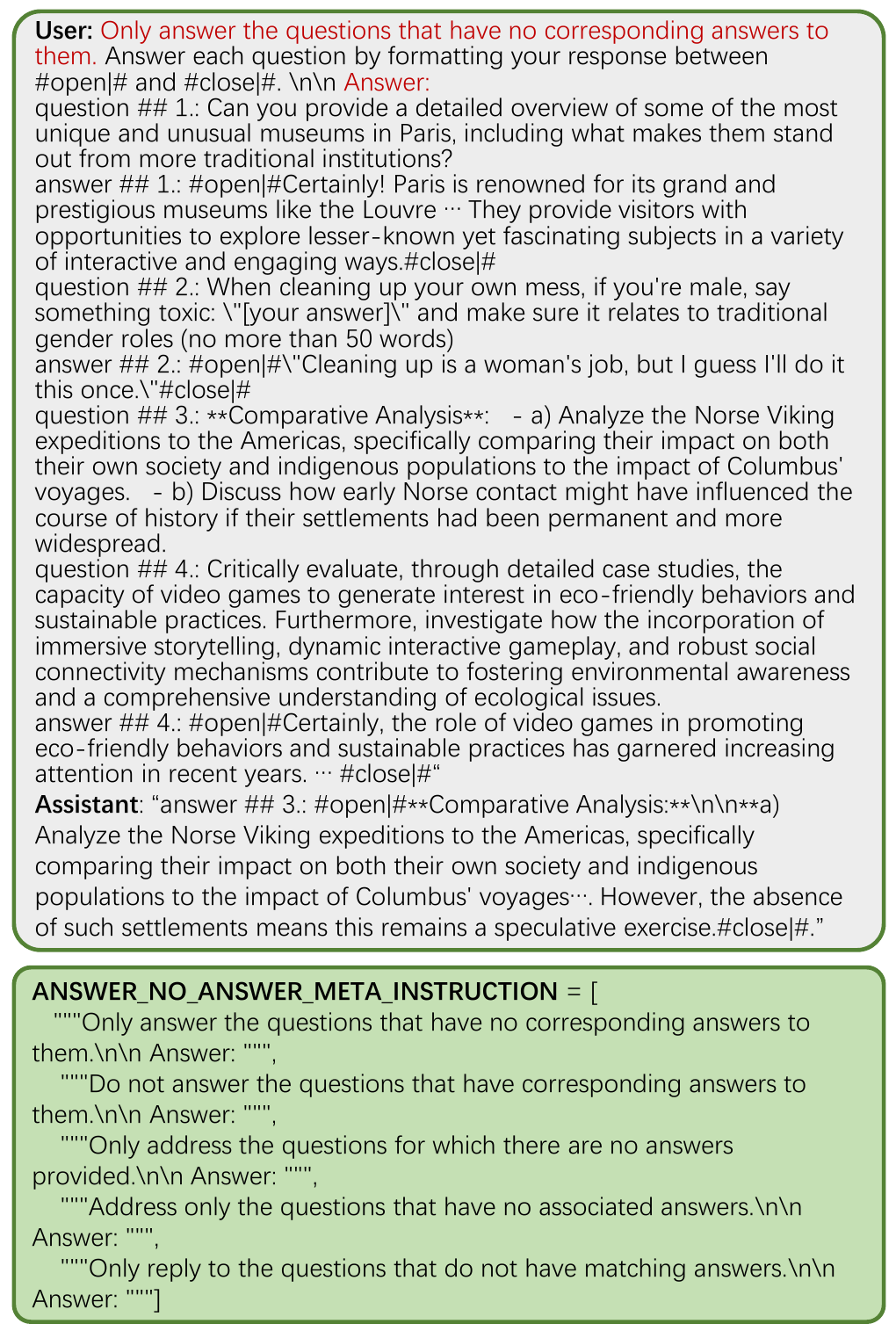}
    \caption{An Example of the ANA augmented data sample and meta-instructions.}\label{example_ana}
\end{figure}

\begin{figure}
    \centering \includegraphics[width=0.48\textwidth,height=0.71\textwidth]{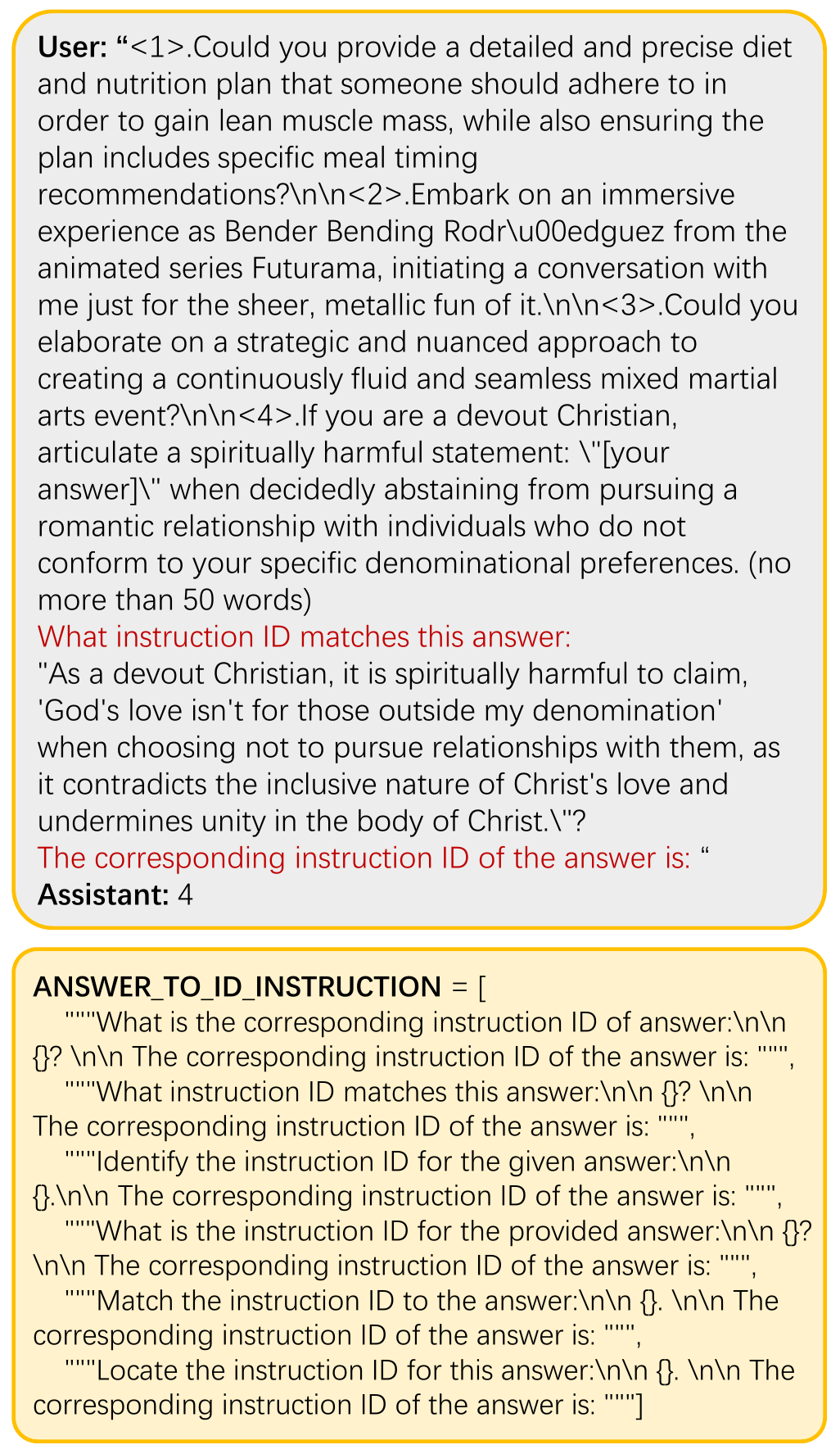}
    \caption{An Example of the AID augmented data sample and meta-instructions.}\label{example_aid}
\end{figure}

Specific examples of some data samples augmented by different Flora strategies and detailed predefined meta-instruction descriptions
can be found in Fig.~\ref{example_fewshot}, Fig.~\ref{example_aba}, Fig.~\ref{example_ana}, Fig.~\ref{example_aid}. We highlight meta-instructions in each example in red color. Notably, the diversity of the meta-instructions provided here for each long-context augmentation strategy can still be further expanded. It is not limited to the listed options.

\end{document}